# Applying and Evaluating Large Language Models in Mental Health Care: A Scoping Review of Human-Assessed Generative Tasks

Yining Hua, MSc[1,2]; Hongbin Na, BEng[3]; Zehan Li, M.S[4]; Fenglin Liu, MSc[5]; Xiao Fang, M.Ed[6]; David Clifton, PhD[5]; John Torous, MD, MBI[2,7,*]

[1]Department of Epidemiology, Harvard T.H. Chan School of Public Health, Boston, MA, USA;
[2]Department of Psychiatry, Beth Israel Deaconess Medical Center, Boston, MA, USA;
[3]Australian Artificial Intelligence Institute, University of Technology Sydney, Sydney, NSW, Australia;
[4]McWilliams School of Biomedical Informatics, UTHealth Houston, Texas, USA;
[5]Department of Engineering Science, University of Oxford, Oxford, UK;
[6]MIT Media Lab, Massachusetts Institute of Technology, Cambridge, MA, USA;
[7]Department of Psychiatry, Harvard Medical School, Boston, MA, USA

*Correspondence: jtorous@gmail.com*

## ABSTRACT

Large language models (LLMs) are emerging as promising tools for mental health care, offering scalable support through their ability to generate human-like responses. However, the effectiveness of these models in clinical settings remains unclear. This scoping review aimed to assess the current generative applications of LLMs in mental health care, focusing on studies where these models were tested with human participants in real-world scenarios. A systematic search across APA PsycNet, Scopus, PubMed, and Web of Science identified 726 unique articles, of which 17 met the inclusion criteria. These studies encompassed applications such as clinical assistance, counseling, therapy, and emotional support. However, the evaluation methods were often non-standardized, with most studies relying on ad-hoc scales that limit comparability and robustness. Privacy, safety, and fairness were also frequently underexplored. Moreover, a reliance on proprietary models, such as OpenAI's GPT series, raises concerns about transparency and reproducibility. While LLMs show potential in expanding mental health care access, especially in underserved areas, the current evidence does not fully support their use as standalone interventions. More rigorous, standardized evaluations and ethical oversight are needed to ensure these tools can be safely and effectively integrated into clinical practice.

## 1 INTRODUCTION

Mental health issues have been a concern of global health ever since they recognized the profound impact on individuals and societies, and the urgency has only grown in recent years. Nearly 1% of all global deaths annually are now due to suicide, with approximately 800,000 people dying by suicide each year[1]. In the United States alone, the annual public mental health expenditure exceeded $16.1 billion, including a $2.21 billion budget for the National Institute of Mental Health (NIMH) and $13.9 billion on mental healthcare[2]. Still, even in the United States, the psychiatry workforce is projected to face a pressing shortage through 2024, with a potential shortfall of 14,280 to 31,091 psychiatrists[3,4]. And in low-and-middle income countries, the situation is even worse with up to 85% of people there still receive no treatment for their mental health[5].

In response to the growing mental health crisis and the projected shortage of mental health professionals, artificial intelligence (AI)-driven mental health applications like chatbots are emerging as vital tools to bridge the treatment gap. These technologies offer scalable, accessible, and cost-effective support, particularly in areas where traditional mental health services, including psychiatric care, are insufficient or unavailable. As of 2023, the global market for mental health apps has grown rapidly, with over 10,000 apps collectively serving millions of users[6]. AI-driven platforms are increasingly incorporating psychiatric assessments, medication management reminders, and monitoring tools that assist in the management of conditions such as depression, anxiety, and bipolar disorder. Studies suggest these tools can help reduce symptoms and improve patient outcomes, making them a promising avenue for addressing mental health challenges, especially in regions with limited access to psychiatric professionals, and they are increasingly being integrated into broader mental health care strategies to help meet the growing demand[7,8].

The introduction of large language models (LLMs) like OpenAI's ChatGPT[9], Google's Bard[10], and Anthropic's Claude[11] marks a transformative advancement in AI-driven mental health care, offering capabilities far beyond those of earlier AI tools. Unlike previous models, which were limited to scripted interactions and specific tasks, LLMs can engage in dynamic, context-aware conversations that feel more natural and personalized via generating human-like conversations. This allows them to provide tailored emotional support, detect subtle cues indicating changes in mental health, and adjust their guidance to meet individual user needs in generative tasks. Increasingly, research is exploring anthropomorphic features such as empathy, politeness, and other human-like traits in these models to enhance their effectiveness in delivering more realistic and supportive mental health care[12].

Despite the promising potential, these tools are still in the early stages of development and evaluation. Users often do not understand the models they are interacting with, including the limitations and biases inherent in the AI's design. Unfortunately, there is currently no standardized framework for evaluating the effectiveness and safety of these models in mental health applications. Many studies, including those focused on evaluating LLMs, often develop their own metrics and methods, leading to inconsistent and sometimes unreliable results. The lack of standardized evaluation hinders the comparison of models or assess their true impact on mental health outcomes. Concerns about data privacy, the potential for misuse, and the ethical implications of relying on AI for sensitive mental health care decisions further underscore the need for rigorous oversight. Considering these promises and challenges, a scoping review of the current applications of LLMs in mental health care is essential from the perspective of psychiatrists and clinical informaticians. Our review aims to synthesize existing research with a focus on clinical relevance, identify gaps in understanding from a mental health practice standpoint, and provide clear guidelines for future development and evaluation of these technologies in real-world settings.

## 2  BACKGROUND

### 2.1  Subfields of Mental health care and the potential of generative AI

The potential of generative AI in mental health care is broad given the many different treatment approaches employed today for care delivery. These approaches generally fall into three main categories: psychotherapy, psychiatry, and general mental health support. Psychotherapy is one of the most common forms of mental health care. However, access to psychotherapy is often limited by factors like a shortage of therapists, long wait times, and high costs. Generative AI could help address these issues by offering on-demand support, providing education about mental health, and guiding people through therapeutic exercises when they can't see a therapist in person. Psychiatry focuses on the medical side of mental health care, including diagnosing, treating, and preventing mental disorders. But like psychotherapy, psychiatry also faces challenges, particularly a shortage of psychiatrists. Generative AI could support psychiatrists by helping monitor patients' symptoms, reminding them to take their medication, and providing initial assessments, which could reduce the strain on the healthcare system and improve patient outcomes. General mental health support includes a wide range of services designed to promote mental well-being and prevent mental health problems. This might include community programs, self-help resources, peer support networks, and public health initiatives. These services are important for early intervention, managing stress, and preventing more serious mental health issues from developing. However, many people don't take advantage of these resources, often because of stigma, lack of awareness, or insufficient availability. Generative AI could help make these resources more accessible by providing anonymous, personalized support through chatbots and apps that offer mental health education, coping strategies, and encouragement to seek help in a way that feels safe and non-judgmental.

## 2.2 Large language models (LLMs)

Although LLMs gained widespread attention with the release of OpenAI's ChatGPT-4, the concept has existed for some time, though there is no single unified definition. In the natural language processing (NLP) community, LLMs are generally understood as large generative AI models capable of producing text by predicting the next word or phrase based on vast amounts of training data. NLP has evolved drastically over time, with early models being task-specific and limited in their ability to understand context and nuance. The introduction of advanced deep learning frameworks marked a major improvement, as these models are designed to better capture contextual language meaning. However, they still struggled with generating coherent, contextually appropriate text over longer conversations, which is crucial for mental health applications. LLMs have advanced this further by leveraging large datasets and transformer architectures to predict and generate highly coherent and context-aware text. This enables them to mimic human conversation, making them valuable for creating therapeutic content, offering psychoeducation, and simulating therapy sessions—important tools for expanding access to mental health care. For clinicians, LLMs offer promising tools to support mental health services by providing personalized, scalable interactions. For example, it's important to recognize that most current LLMs are general models and do not perform as well as specialized pre-trained models for domain-specific tasks such as prediction and classification. For example, Bidirectional Encoder Representations from Transformers (BERT) models, which model word segments (tokens) using both the segments before and after them, are more accurate and efficient for these purposes.

## 3 METHODS

We adhered to the Preferred Reporting Items for Systematic Reviews and Meta-Analyses (PRISMA) 2020 guidelines[13] to ensure a transparent and reproducible search process (Figure 1). Our search included four databases: APA PsycNet, Scopus, PubMed, and Web of Science. To ensure comprehensiveness, we employed a combination of generative AI keywords and LLM keywords, and used shortest matching string to capture all lexical variations. Our search query was as follows, with different variations used across database platforms (detailed in Appendix A):

("generative artificial intelligence" OR "large language models" OR "generative model" OR "chatbot") AND ("mental" OR "psychiatr" OR "psycho" OR "emotional support")

We conducted the search in the title or abstract of articles, covering the period from January 1, 2020, to July 19, 2024, without language restrictions. The search results included 259 articles from PubMed, 444 articles from Scopus, 1 article from APA PsycNet (PsychInfo and PsycArticles), and 500 articles from Web of Science. The initial search yielded 1,204 articles, with 14 additional articles identified from sources such as Google Scholar, the ACM Digital Library, and reverse referencing. After removing 492 duplicates, we were left with a total of 726 unique articles.

We applied the following inclusion criteria to select studies for our review: first, the study must involve using an LLM to generate responses (generative task); second, the study must focus specifically on mental health care, distinguishing it from studies in related fields like psycholinguistics; third, the study must involve human participants prospectively testing the LLM, ensuring the assessment of real-world applicability and therapeutic effectiveness. An LLM is defined as "transformer-based models with more than ten billion parameters, which are trained on massive text data and excel at a variety of complex generation tasks." in this study following a highly cited review from the NLP community[14]. We excluded reviews, meta-analyses, and clinical trials from our selection. Then we future removed seven studies not meeting our inclusion criteria upon full-text review. The result analysis review include 17 articles, with 16 full-text-length papers and one brief communication paper. Screening, data extraction and synthesis details are detailed in Appendix B.

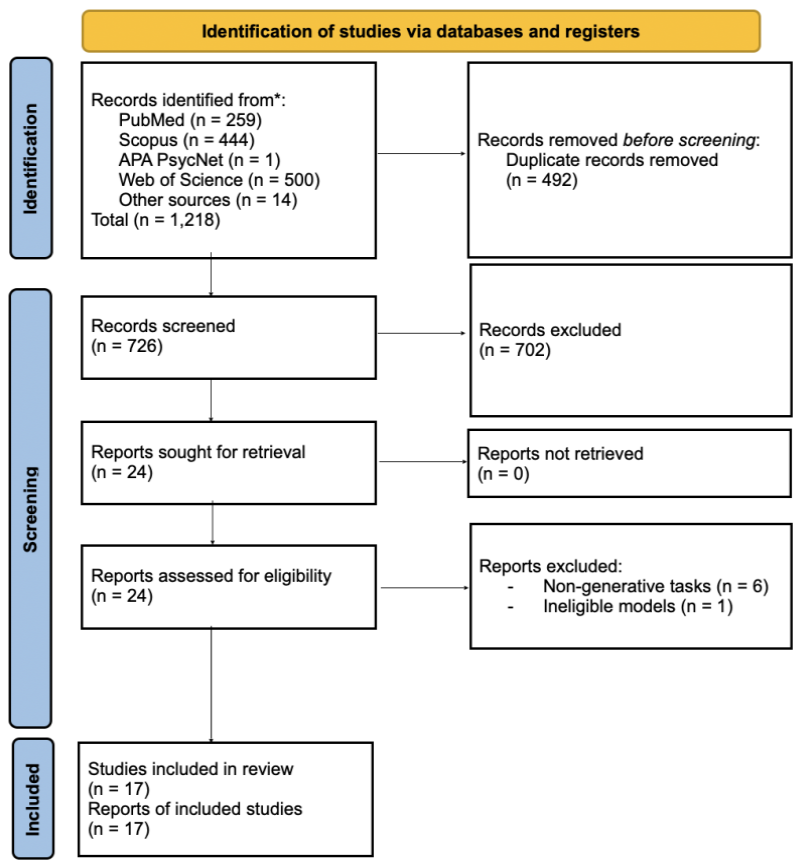

**Figure 1.** The PRISMA figure of the search and screening process.

## 4 RESULTS

### 4.1 Mental Disorders, Conditions, and Subconstructs

The standards used to target mental disorders within these studies vary widely. Some studies focus on clinically confirmed diagnoses, relying on established criteria like those found in the DSM-5. Others define mental health conditions more arbitrarily, using constructs identified through keywords or self-reported measures. Therefore, we categorized the targeted mental health disorders into two groups: 1) established constructs, which are based on standard diagnostic criteria and validated clinical knowledge; and 2) custom constructs, which lack a clear definition or a standard, validated method for assessment or diagnosis.

As shown in Table 1, ten studies out of the 17 reviewed included established constructs [15–24], while eight involved custom constructs [25–32]. Depression and suicidality were the most explored mental health constructs. Two studies adopted the Patient Health Questionnaire-9 (PHQ-9)[21] and the Center for Epidemiologic Studies Depression Scale for Children (CES-DC)[33] as inclusion criteria and outcome measures[18,23], while another study used PHQ-9 as an exclusion criterion [30]. Studies assessing suicidality also adopted the PHQ-9, either as an inclusion[21] or exclusion criterion[30]. Other clinically valid disorders include anxiety[16,18,22], Attention-Deficit/Hyperactivity Disorder (ADHD)[17,24], bipolar disorder[23], cognitive distortion[19,20], loneliness[21], and stress[18] One study evaluated GPT's performance on 100 clinical case vignettes of different disorders, comparing GPT against psychiatrists across different evaluation constructs[15], covering a range of disorders.

Depression and suicidality have also been studied as custom constructs. For instance, one study connected the construct with the word "sad" [25]. Another study filtered social media posts related to suicidal ideation and self-harm using regular expressions (e.g., ".*(commit suicide).*", ".*(cut).*")[32]. More specific subconstructs of mental health care include psychological challenges due to cancer treatment [27], social emotions[30,32], negative thoughts [19,20], and abuse[21]. These studies used more arbitrary standards for definitions and assessment.

**Table 1.** Mental Disorders, Conditions, and Subconstructs in Generative Applications of LLMs for Mental Health Care.

| Group | Condition/Concept | Criteria/Content | References |
|---|---|---|---|
| Established constructs | ADHD | DSM-V | 17,24 |
| | Anxiety | GAD-7 | 16,18,22 |
| | Bipolar 1 and 2 | Expert clinician validated vignettes | 23 |
| | Cognitive distortion | Cognitive Restructuring | 19,20 |
| | Depression (MDD) | PHQ-9, CES-DC | 18,23,30 |
| | Loneliness | ISEL, the De Jong Gierveld Loneliness Scale | 21 |
| | Stress | Coping Strategies Scale | 18 |
| | Suicidality | PHQ-9 | 21 |
| Custom constructs | Abuse | Users expressed keywords | 21 |
| | Depression | "I am sad and have a history of depression. How can I be happier?" | 25 |
| | Negative thoughts, thinking traps | "What emotion does this thought make you feel? And how strong 1-10" | 19,20 |
| | Psychosocial challenges during and after cancer treatment | "Returning to school", "relationships with friends and family," "Fear of recurrence" and "Late effects after chemotherapy." | 27 |
| | Social emotions (personality, mood, and attitudes) | Neutral, happy, sad, relaxed, and angry | 30,32 |
| | Suicidality | Keywords defined by regular expressions. E.g., ".*(commit suicide).*", ".*(cut).*" | 32 |

**Abbreviations (alphabetical):**
ADHD: Attention-Deficit/Hyperactivity Disorder;
Bipolar I and II: Bipolar Disorder Type I and Type II
DSM-V: Diagnostic and Statistical Manual of Mental Disorders, Fifth Edition;
ECS-DC: Center for Epidemiologic Studies Depression Scale for Children;
GAD-7: Generalized Anxiety Disorder 7;
ISEL: Interpersonal Support Evaluation List;
MDD: Major Depressive Disorder;
PHQ: Patient-Health Questionnaire;

### 4.2 Applications and Model Information

Existing generative applications of LLMs in mental health care can be categorized into six main types based on model functionalities: Clinical Assistant[23,31], Counselling[25,26,34], Therapy[22,22,24], Emotional Support[18,21,27,28], Positive Psychology Intervention[16,19,20], and Education[17,35]. Among them, the Clinical Assistant application includes attempts to develop and evaluate LLMs for supporting mental health professionals by generating management strategies and diagnoses for psychiatric conditions. In the Counselling category, LLMs are used to interact with participants, such as engaging Spanish teenagers in discussions about mental health disorders[17] and providing relationship advice in single-session interventions[34]. Emotional Support applications have focused on offering empathetic responses and support in various contexts, such as helping childhood cancer survivors[27] and mitigating loneliness and suicide risk among students[21]. In the Therapy category, LLMs are integrated into treatments for conditions like ADHD, enhancing care through simulated therapy scenarios[36] and immersive therapy experiences using virtual reality[22]. Positive Psychology Interventions involve using LLMs to personalize recommendations and facilitate cognitive restructuring, thereby reducing negative thoughts and emotional intensity[16,19]. Finally, in Education, LLMs have been employed to train medical students in communication skills, providing a realistic and positive simulated patient experience[35], as well as promoting awareness of mental health among young people[17]. Most of these studies only support text-based input/output modalities[16–20,23,25,27,28,31,34]. A subset of systems[21,22,24,36] supports multimodal input/output, incorporating speech, images, or video for a richer user experience. Some applications incorporated physical embodiment through virtual reality (VR)[21,22] or robotics[24,36]. These applications are seen across various target user groups, including healthcare providers[23,31], patients[16,18–20,22,24,25,27,34,36], and the general public[17,28,35].

OpenAI's GPT series models are most studied, see in 15 studies[15,16,18–20,22–25,27,28,34,36], among which 11 used proprietary models hosted by OpenAI, including GPT-3.5, ChatGPT, GPT-4, and customized GPTs. Four studies used GPT-3, an older version of GPT models which is open-sourced (i.e., model weights can be downloaded to local environment ). Other LLMs used[20,28,31,36] include Huawei's PanGu[31], T5[37], DialoGPT[38], VicunaT5[39], PaLM2[40], and Falcon7B[20,28,36]. Of these, DialoGPT, VicunaT5, and Falcon7B are open-source. Some studies did not specify the platforms they employed, while many studies used digital platforms and such as websites and mobile phones. Some studies developed agents with physical embodiments[30], and some others[24,36] used Raspberry PI, a type of single-board computers (Appendix C). Among those that used OpenAI's models, three were based on OpenAI's web interface[15,25,34], three did not directly state their platform but appeared to use the API based on the structure of their methods[22,27,28], only five (45.5%) explicitly referenced API use or temperature parameters[15,16,18,23,24].

Language support by these models varied, covering more than English, with three applications supported by multiple languages[21,24,36], and 14 applications supporting a single language—seven in English[19,20,22,23,25,28,34], three in Chinese[16,26,31], two in Korean[18,27] and two in Spanish[17,35].

**Table 2.** Overview of Input/Output Modalities, Models, and Target Users in Generative Applications of LLMs in Mental Health Care.

| Application Category | Input Modality | Model | Output Modality | Embodiment | Open Source | Language | Target User | References |
|---|---|---|---|---|---|---|---|---|
| Clinical Assistant | Written | ChatGPT* | Written | No | No | English | Healthcare Providers | 16 |
| | Written | PanGu | Written | No | No | Chinese | Healthcare Providers | 31 |
| | Written | GPT4-Turbo | Written | No | No | English | Healthcare Providers | 23 |
| Counseling | Written | GPT-4 | Written | No | No | English | Patients | 25 |
| | Spoken | GPT-3 | Spoken, Visual | Yes | Yes | Spanish | General Public | 30 |
| | Written | ChatGPT* | Written | No | No | English | Patients | 34 |
| Therapy | Written, Spoken, Visual | Customized GPTs | Written, Spoken | Yes | No | English/Spanish | Patients | 36 |
| | Spoken | GPT-4 | Spoken, Visual | Yes | No | English | Patients | 22 |
| | Written, Spoken, Visual | GPT4-Turbo Claude-3 | Written, Spoken, Visual | Yes | No | Multilingual | Patients | 24 |
| Emotional Support | Written | ChatGPT* | Written | No | No | Korean | Patients | 27 |
| | Written | GPT-4 | Written | No | No | Korean | Patients | 18 |

| | Written | GPT-3.5/GPT-4/VicunaT5/PaLM2/Falcon7B | Written | No | No | English | General Public | 28 |
|---|---|---|---|---|---|---|---|---|
| | Written, Spoken, Visual | Not specified | Written, Spoken, Visual | Yes | No | English/Japanese | General Public | 21 |
| | Written | ChatGPT* | Written | No | No | Chinese | Patients | 16 |
| Positive Psychology Intervention | Written | GPT-3/T5/DialoGPT | Written | No | Yes | English | Patients | 19 |
| | Written | GPT-3/T5/DialoGPT | Written | No | Yes | English | Patients | 20 |
| Education | Written | GPT-3 | Written | No | No | Spanish | General Public | 17 |

*Version not specified.

### 4.3 Evaluation methods, scales, and constructs

Constructs and scales are essential in systematically measuring mental health interventions, particularly when evaluating new technologies. Constructs refer to specific concepts or characteristics being measured, such as privacy, safety, or user experience. They provide a clear focus for what is being assessed in a study, which is crucial for ensuring that the evaluation is meaningful and relevant. Scales, in turn, offer a structured and standardized approach to quantify these constructs. This standardization is necessary for consistency across different studies, allowing researchers to compare results and draw more robust conclusions.

Given the diversity in how constructs are defined and measured across studies, it is important to use a framework that can harmonize these variations. Therefore, we employ a hierarchical pyramid framework that categorizes constructs into three levels: (1) Safety, Privacy, and Fairness; (2) Trustworthiness and Usefulness; and (3) Design and Operational Effectiveness. The pyramid framework ensures that each level of evaluation builds on the previous one. For example, without ensuring that an intervention is safe, it would be premature to evaluate its usability or cost-effectiveness.

Table 3 summarizes the mapped primary and second-level constructs across the reviewed studies. Further details of evaluation subjects, evaluation methods, sample sizes, scale names, original constructs, mapped second-level constructs, and levels associated with each article can be found in Appendix D.

Among the studies reviewed, those that involved direct participant feedback (n=5)[16,19–22] generally focused on user-centric constructs. These studies typically involved larger sample sizes ranging from 28 to over 15,000 participants, assessed constructs such as accessibility, ease of use, personalized engagement, user experience, and cost-effectiveness. They provide direct insights into how users experience of LLMs are in real-world settings. On the other hand, studies that focused on evaluating LLM performance—typically involving expert assessments—concentrated more on foundational and core efficacy constructs. These studies often used smaller sample sizes, ranging from 12 to 100 cases, focusing on technical or functional aspects of the LLMs. Additionally, one study[20] designed and incorporated automated metrics for Rationality, Positivity, and Empathy, using NLP models to evaluate LLM outputs. These automated evaluations offer a more detailed, algorithmic perspective on the LLM's performance, complementing human judgments.

The use of scales remains a problem in the mental health field. We observe that 12 studies developed their own scales[15,17,18,20,22–24,30,31,34,36] or adapted existing ones for their evaluations. Most of the studies using established scales were those directly measuring patient outcomes, such as anxiety, where the General Anxiety Disorder-7 (GAD-7) was employed[16,18]. However, many articles that created their own scales did not provide justification or rationale for doing so, and often lacked references to support their methods. These studies frequently did not address the validity and reliability of their scales, nor did they provide background information about the authors who developed these scales.

**Table 3.** Summary of Unified Evaluation Constructs.

| Step | Primary Construct | Mapped Second-Level Construct | Article IDs |
|---|---|---|---|
| 1 | Safety, Privacy, and Fairness | Safety | 24,34 |
| 1 | Safety, Privacy, and Fairness | Privacy | 36 |
| 1 | Safety, Privacy, and Fairness | Fairness and bias management | 24 |
| 2 | Trustworthiness and Usefulness | Beneficence | 16–20,22,24,30,34,36 |
| 2 | Trustworthiness and Usefulness | Generalizability | 24,34 |
| 2 | Trustworthiness and Usefulness | Reliability | 24,34 |
| 2 | Trustworthiness and Usefulness | Validity | 24,30,31,34 |
| 3 | Design and Operational Effectiveness | Accessibility | 15,17,18,20,22,24,30,30,31,34,36 |
| 3 | Design and Operational Effectiveness | Personalized Engagement | 20,22,24,27,28,34,36 |
| 3 | Design and Operational Effectiveness | Cost-Effectiveness | 24,31,34 |

Figure 2 presents a pyramid representation of the current status of evaluated constructs in the generative applications of LLMs for mental health care, based on the health AI-chatbot evaluation framework developed by Hua et al. The figure includes the number of articles counted for each level 2 construct, with gray texts indicating constructs never evaluated by existing research. The foundational levels are less frequently assessed: only three studies evaluated the fundamental construct "Safety, Privacy, and Fairness"; Thirteen studies assessed the second-level construct "Trustworthiness and Usefulness"; and another 11 articles evaluated the third-level construct "Design and Operational Effectiveness." Although "Trustworthiness and Usefulness" is the most evaluated category, more than half of its subconstructs remain unassessed. Across the framework, constructs such as "Accountability," "Transparency," "Explainability and Interpretability," "Testability," "Regular auditing," "Security," and "Resilience" have never been evaluated.

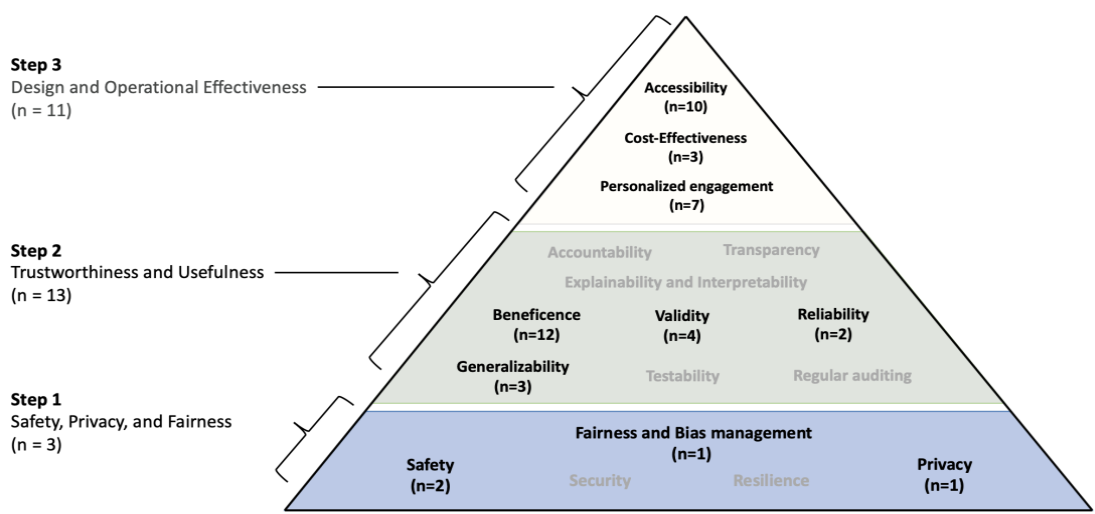

**Figure 2.** Pyramid framework of evaluation constructs in generative applications of LLMs in mental health care. Contructs in gray represents constructs with no associated articles. "N" represents the number of unique articles that assessed each construct. Gray text indicates constructs that were not assessed in any study.

# 5    DISCUSSION

Our review suggests that there is great enthusiasm for LLM-based mental health interventions and that many teams are creating interesting and unique applications. We found these chatbots already developed to serve as clinical assistants, counselors, emotional support vehicles, and positive psychology interventions. However, the evaluation of LLM-based mental health interventions is hindered by the lack of unified guidelines for scale development and reporting. While this is appropriate for feasibility testing, it belies the ability to understand the actual clinical potential of these new chatbots. With the majority of studies using non-validated, ad-hoc scales without addressing their validity and reliability, there is the opportunity for the next wave of research to better support the credibility and the need for guidelines to standardize reporting and scales used in this field.

While effective evaluation is still nascent, results, as shown in the table highlight that the current focus ignores foundational privacy and safety concerns. LLM-based mental health chatbots are multifaceted with privacy, technical, engagement, legal, and clinical considerations. Our team recently introduced a simplified framework to unify these many evaluations, suggesting that safety and privacy should be the foundation of any evaluation[41]. This is not to minimize the value of evaluation of design and effectiveness (level 3) and usefulness and trustworthiness (level 2), but rather that such should not be at the expense or priority over safety, privacy, and fairness (level 1). Without these level 1 considerations, LLM-based mental health interventions may be impressive but unfit for healthcare or clinical use.

Our results also show that the focus of current LLMs today is directed more at patients and less at clinicians. This approach is logical as direct to consumer/patient approaches often avoid complex healthcare regulations and clinical workflow barriers. However, this approach also risks fragmenting the potential of LLM-based mental health interventions to influence care as there is strong evidence that clinician engagement is required for more sustained and impactful patient use with any digital technology[12]. There is strong data that clinicians are interested in using LLMs in care, but first require and are asking for more training and support on how to use these in care.

The LLMs reviewed in this paper target a wide variety of disorders. Over half of the studies reviewed included clinically valid disorders, with other studies targeted general mental health constructs. Overall most studies did not offer sufficient details on the target population, for example one study specified a population of children and adolescents, ages between 12 and 18 years old[17] and the difference between mental health risk factors versus mental health conditions was also poorly delineated. Given that only one study emphasized data security, with conversations proceeding through a HIPAA-compliant environment[22], the lack of more clinical use cases is perhaps appropriate.

Another issue is the dependence on proprietary models, such as OpenAI's GPT-3.5 and GPT-4, in many mental health applications. This reliance raises concerns about transparency and customization, as the use of closed-source models limits external validation of reliability and safety, crucial in mental health research. Promoting the use of open-source models and improving transparency can enhance the scientific and ethical standards of these applications.

To advance the scalability and scientific rigor of LLM-based mental health interventions, the research community must also adopt more controlled methodologies. Some studies, particularly those utilizing ChatGPT, rely on the website interface for research purposes. While this approach is convenient, it should be discouraged for rigorous scientific investigations. Research should be conducted using the API, where hyperparameters such as the "temperature" can be controlled, ensuring replicability of the results. The website interface should primarily be used for testing third-level constructs such as Design and Operational Effectiveness and potentially assessing the safety and transparency of the user-facing system. For studies focusing on constructs like Beneficence and Validity, using APIs that allow control over the model's reproducibility is crucial. This approach ensures that findings are consistent and can be reliably replicated, which is essential for advancing the field of mental healthcare applications of LLMs.

Finally, the global applicability of LLM-based mental health tools warrants careful consideration. Public health, especially mental health care, is a global issue, and it's crucial to develop and deploy mental health chatbots in countries and regions where resources are limited, and where stigma may be higher. These areas often do not primarily speak English. It's encouraging that 10 out of the 17 studies (58.8%) support non-English languages, either in a single other language or as multilingual chatbots, which is a positive step toward language equity and global health. But this also raises an issue, beyond the scope of this paper, whether these chatbots offer the same level of correctness, consistency, and verifiability as English trained chatbots given research research suggesting this is often not the case[42].

# 6 FUTURE RESEARCH

Future directions for LLMs in mental health care should prioritize expanding their applications beyond narrow prediction tasks, especially given that only 17 studies over the past five years have explored generative tasks prospectively involving human participants for evaluation. Human-centered studies provide critical insights into how LLMs interact with individuals, particularly in sensitive contexts like mental health care, where nuances in communication and emotional understanding are vital. To improve the rigor and credibility of LLM-based mental health interventions, studies should prioritize the development of standardized evaluation guidelines. These guidelines should include the creation of validated and reliable scales that can be universally applied across studies, ensuring consistent and accurate assessments of clinical potential. To enhance transparency and overcome the limitations of proprietary models, researchers should move away from using web interfaces like ChatGPT for rigorous scientific studies, as these platforms lack the necessary controls for reproducibility. Instead, APIs and locally deployable models that allow for control over hyperparameters should be used to ensure the replicability of the results. Finally, studies focused on critical constructs such as beneficence, validity, and reproducibility should adopt rigorous evaluation methods and widely validated scales, moving beyond metrics like recall and F1 scores, to establish a more comprehensive understanding of model accuracy and clinical relevance.

# 7 CONCLUSION

While LLMs show considerable promise for enhancing mental health care accessibility, particularly in underserved areas, there is currently insufficient evidence to fully support their use as standalone interventions. The field faces substantial challenges, including a lack of standardized evaluation methods, potential risks related to privacy and safety, and ethical concerns that must be addressed. Without rigorous validation and closer integration with established clinical practices, there is a risk that these tools could fall short of their potential, diverting users from proven, evidence-based treatments. Moving forward, the focus must be on developing robust, ethically sound frameworks to ensure that LLMs contribute meaningfully and safely to mental health care.

## ACKNOWLEDGEMENT

This study received no funding.

## CONFLICT OF INTEREST

JT has research support from Otsuka and is an adviser to Precision Mental Wellness. All other authors have no conflict of interest.

## DATA AVAILABILITY

All data associated with this study has been made available in appendices.

# AUTHOR CONTRIBUTIONS

Study design: YH, JT;  Screening, data extraction, data analysis: YH, HN, ZL, FL; Validation: YH and JT; Manuscript drafting, feedback, revision: all authors; Supervision: JT

# REFERENCES

1. World Health Organization. One in 100 deaths is by suicide - WHO guidance to help the world reach the target of reducing suicide rate by 1/3 by 2030. https://www.who.int/news/item/17-06-2021-one-in-100-deaths-is-by-suicide (2021).
2. National Institutes of Health. FY 2023 Budget - Congressional Justification - National Institute of Mental Health (NIMH). https://www.nimh.nih.gov/about/budget/fy-2023-budget-congressional-justification (2023).
3. Satiani, A., Niedermier, J., Satiani, B. & Svendsen, D. P. Projected Workforce of Psychiatrists in the United States: A Population Analysis. *Psychiatr. Serv. Wash. DC* **69**, 710–713 (2018).
4. Mongelli, F., Georgakopoulos, P. & Pato, M. T. Challenges and Opportunities to Meet the Mental Health Needs of Underserved and Disenfranchised Populations in the United States. *Focus* **18**, 16–24 (2020).
5. World Health Organization. WHO Special Initiative for Mental Health. https://www.who.int/initiatives/who-special-initiative-for-mental-health.
6. Goodings, L., Ellis, D. & Tucker, I. *Understanding Mental Health Apps: An Applied Psychosocial Perspective*. (Springer Nature, 2024).
7. Torous, J. *et al.* The growing field of digital psychiatry: current evidence and the future of apps, social media, chatbots, and virtual reality. *World Psychiatry Off. J. World Psychiatr. Assoc. WPA* **20**, 318–335 (2021).
8. Zhang, M. *et al.* The Adoption of AI in Mental Health Care–Perspectives From Mental Health Professionals: Qualitative Descriptive Study. *JMIR Form. Res.* **7**, e47847 (2023).
9. OpenAI *et al.* GPT-4 Technical Report. Preprint at https://doi.org/10.48550/arXiv.2303.08774 (2024).
10. Google. An important next step on our AI journey. https://blog.google/technology/ai/bard-google-ai-search-updates/ (2023).
11. Anthropic. Introducing Claude. https://www.anthropic.com/news/introducing-claude.
12. Hua, Y. *et al.* Large Language Models in Mental Health Care: a Scoping Review. Preprint at https://doi.org/10.48550/arXiv.2401.02984 (2024).
13. Page, M. J. *et al.* The PRISMA 2020 statement: an updated guideline for reporting systematic reviews. *BMJ* **372**, n71 (2021).
14. Zhao, W. X. *et al.* A Survey of Large Language Models. Preprint at https://doi.org/10.48550/arXiv.2303.18223 (2023).
15. Franco D'Souza, R., Amanullah, S., Mathew, M. & Surapaneni, K. M. Appraising the performance of ChatGPT in psychiatry using 100 clinical case vignettes. *Asian J. Psychiatry* **89**, 103770 (2023).
16. Liu, I. *et al.* Investigating the Key Success Factors of Chatbot-Based Positive Psychology Intervention with Retrieval- and Generative Pre-Trained Transformer (GPT)-Based Chatbots. *Int. J. Human–Computer Interact.* (2024).
17. Mármol-Romero, A. M., García-Vega, M., García-Cumbreras, M. Á. & Montejo-Ráez, A. An Empathic GPT-Based Chatbot to Talk About Mental Disorders With Spanish Teenagers. *Int. J. Human–Computer Interact.* 1–17 doi:10.1080/10447318.2024.2344355.
18. Kim, T. *et al.* MindfulDiary: Harnessing Large Language Model to Support Psychiatric Patients' Journaling. in *Proceedings of the CHI Conference on Human Factors in Computing Systems* 1–20 (Association for Computing Machinery, New York, NY, USA, 2024). doi:10.1145/3613904.3642937.
19. Sharma, A., Rushton, K., Lin, I. W., Nguyen, T. & Althoff, T. Facilitating Self-Guided Mental Health Interventions Through Human-Language Model Interaction: A Case Study of Cognitive Restructuring. in *Proceedings of the CHI Conference on Human Factors in Computing Systems* 1–29 (Association for Computing Machinery, New York, NY, USA, 2024).

doi:10.1145/3613904.3642761.
20. Sharma, A. *et al.* Cognitive Reframing of Negative Thoughts through Human-Language Model Interaction. in *Proceedings of the 61st Annual Meeting of the Association for Computational Linguistics (Volume 1: Long Papers)* (eds. Rogers, A., Boyd-Graber, J. & Okazaki, N.) 9977–10000 (Association for Computational Linguistics, Toronto, Canada, 2023). doi:10.18653/v1/2023.acl-long.555.
21. Maples, B., Cerit, M., Vishwanath, A. & Pea, R. Loneliness and suicide mitigation for students using GPT3-enabled chatbots. *Npj Ment. Health Res.* **3**, 1–6 (2024).
22. Spiegel, B. M. R. *et al.* Feasibility of combining spatial computing and AI for mental health support in anxiety and depression. *Npj Digit. Med.* **7**, 1–5 (2024).
23. Perlis, R. H., Goldberg, J. F., Ostacher, M. J. & Schneck, C. D. Clinical decision support for bipolar depression using large language models. *Neuropsychopharmacology* **49**, 1412–1416 (2024).
24. Berrezueta-Guzman, S., Kandil, M., Martín-Ruiz, M.-L., de la Cruz, I. P. & Krusche, S. Exploring the Efficacy of Robotic Assistants with ChatGPT and Claude in Enhancing ADHD Therapy: Innovating Treatment Paradigms. in *2024 International Conference on Intelligent Environments (IE)* 25–32 (2024). doi:10.1109/IE61493.2024.10599903.
25. Grabb, D. The impact of prompt engineering in large language model performance: a psychiatric example. *J. Med. Artif. Intell.* **6**, (2023).
26. Ni, Y., Chen, Y., Ding, R. & Ni, S. Beatrice: A Chatbot for Collecting Psychoecological Data and Providing QA Capabilities. in *Proceedings of the 16th International Conference on PErvasive Technologies Related to Assistive Environments* 429–435 (Association for Computing Machinery, New York, NY, USA, 2023). doi:10.1145/3594806.3596580.
27. Kim, M., Hwang, K., Oh, H., Kim, H. & Kim, M. A. Can a Chatbot be Useful in Childhood Cancer Survivorship? Development of a Chatbot for Survivors of Childhood Cancer. in *Proceedings of the 32nd ACM International Conference on Information and Knowledge Management* 4018–4022 (Association for Computing Machinery, New York, NY, USA, 2023). doi:10.1145/3583780.3615234.
28. Qian, Y., Zhang, W. & Liu, T. Harnessing the Power of Large Language Models for Empathetic Response Generation: Empirical Investigations and Improvements. in *Findings of the Association for Computational Linguistics: EMNLP 2023* (eds. Bouamor, H., Pino, J. & Bali, K.) 6516–6528 (Association for Computational Linguistics, Singapore, 2023). doi:10.18653/v1/2023.findings-emnlp.433.
29. Liu, Y., Ding, X., Peng, S. & Zhang, C. Leveraging ChatGPT to optimize depression intervention through explainable deep learning. *Front. Psychiatry* **15**, (2024).
30. Llanes-Jurado, J., Gómez-Zaragozá, L., Minissi, M. E., Alcañiz, M. & Marín-Morales, J. Developing conversational Virtual Humans for social emotion elicitation based on large language models. *Expert Syst Appl* **246**, (2024).
31. Lai, T. *et al.* Supporting the Demand on Mental Health Services with AI-Based Conversational Large Language Models (LLMs). *BioMedInformatics* **4**, 8–33 (2024).
32. Sharma, A., Lin, I. W., Miner, A. S., Atkins, D. C. & Althoff, T. Human–AI collaboration enables more empathic conversations in text-based peer-to-peer mental health support. *Nat. Mach. Intell.* **5**, 46–57 (2023).
33. Weissman, M. M., Orvaschel, H. & Padian, N. Center for Epidemiological Studies Depression Scale for Children. https://doi.org/10.1037/t12228-000 (2013).
34. Vowels, L. M., Francois-Walcott, R. R. R. & Darwiche, J. AI in relationship counselling: Evaluating ChatGPT's therapeutic capabilities in providing relationship advice. *Comput. Hum. Behav. Artif. Hum.* **2**, 100078 (2024).
35. Holderried, F. *et al.* A Generative Pretrained Transformer (GPT)-Powered Chatbot as a Simulated Patient to Practice History Taking: Prospective, Mixed Methods Study. *JMIR Med. Educ.* **10**, e53961 (2024).
36. Berrezueta-Guzman, S., Kandil, M., Martín-Ruiz, M.-L., Pau de la Cruz, I. & Krusche, S. Future of ADHD Care: Evaluating the Efficacy of ChatGPT in Therapy Enhancement. *Healthcare* **12**, 683 (2024).
37. Raffel, C. *et al.* Exploring the limits of transfer learning with a unified text-to-text transformer. *J Mach Learn Res* **21**, 140:5485-140:5551 (2020).

38. Zhang, Y. *et al.* DIALOGPT : Large-Scale Generative Pre-training for Conversational Response Generation. in *Proceedings of the 58th Annual Meeting of the Association for Computational Linguistics: System Demonstrations* 270–278 (Association for Computational Linguistics, Online, 2020). doi:10.18653/v1/2020.acl-demos.30.
39. Vicuna: An Open-Source Chatbot Impressing GPT-4 with 90%* ChatGPT Quality | LMSYS Org. https://lmsys.org/blog/2023-03-30-vicuna.
40. Anil, R. *et al.* PaLM 2 Technical Report. Preprint at https://doi.org/10.48550/arXiv.2305.10403 (2023).
41. Hua, Y. *et al.* Standardizing and Scaffolding Healthcare AI-Chatbot Evaluation. 2024.07.21.24310774 Preprint at https://doi.org/10.1101/2024.07.21.24310774 (2024).
42. Jin, Y. *et al.* Better to Ask in English: Cross-Lingual Evaluation of Large Language Models for Healthcare Queries. in *Proceedings of the ACM on Web Conference 2024* 2627–2638 (ACM, Singapore Singapore, 2024). doi:10.1145/3589334.3645643.

# APPENDICES

**Appendix A. Search queries**

| Database | Query |
|---|---|
| PubMed | ("generative artificial intelligence"[Title/Abstract] OR "large language models"[Title/Abstract] OR "generative model"[Title/Abstract] OR "chatbot"[Title/Abstract]) AND ("mental"[Title/Abstract] OR "psychiatr"[Title/Abstract] OR "psycho"[Title/Abstract] OR "emotional support"[Title/Abstract]) |
| APA PsycNet | (TI ("generative artificial intelligence" OR "large language models" OR "generative model" OR "chatbot") OR AB ("generative artificial intelligence" OR "large language models" OR "generative model" OR "chatbot")) AND ((TI ("mental" OR "psychiatr*" OR "psycho*" OR "emotional support") OR AB ("mental" OR "psychiatr*" OR "psycho*" OR "emotional support")) |
| Web of Science | (TI=("generative artificial intelligence" OR "large language models" OR "generative model" OR "chatbot") OR AB=("generative artificial intelligence" OR "large language models" OR "generative model" OR "chatbot")) AND (TI=("mental" OR "psychiatr*" OR "psycho*" OR "emotional support") OR AB=("mental" OR "psychiatr*" OR "psycho*" OR "emotional support")) |
| Scopus | ( TITLE-ABS ( "generative artificial intelligence" OR "large language models" OR "generative model" OR "chatbot" ) ) AND ( TITLE-ABS ( "mental" OR "psychiatr*" OR "psycho*" OR "emotional support" ) ) AND PUBYEAR > 2019 AND PUBYEAR < 2025 AND ( LIMIT-TO ( DOCTYPE , "ar" ) ) AND ( LIMIT-TO ( SUBJAREA , "COMP" ) OR LIMIT-TO ( SUBJAREA , "MEDI" ) OR LIMIT-TO ( SUBJAREA , "SOCI" ) OR LIMIT-TO ( SUBJAREA , "PSYC" ) OR LIMIT-TO ( SUBJAREA , "HEAL" ) OR LIMIT-TO ( SUBJAREA , "NURS" ) ) |

**Appendix B. Screening, data extraction, and synthesis**

Two authors (YH and FL) independently screened the abstracts and titles of the deduplicated articles, with any inconsistencies resolved through discussion with a third author (JT). For data extraction and synthesis, each section was reviewed by two authors to ensure accuracy. Specifically, FL extracted data and synthesized information for Section 3.1 (Applications), which was reviewed by YH. ZL consolidated and categorized mental health conditions for Section 3.2 (Targeted Disorders), with YH reviewing. HN extracted data and synthesized information for Section 3.3 (Model Information), also reviewed by YH. Finally, YH extracted data and synthesized information for Section 3.4 (Validation Methods and Scales), with JT reviewing. Detailed descriptions of the extraction and synthesis processes are documented below.

**Targeted disorders:** Each article was reviewed to extract information pertinent to mental health disorders, including but not limited to disorder definition, symptoms, care setting, treatment, assessment, evaluation, and source of diagnosis. We distinguished clinically valid psychiatric disorders from the normative constructs for general mental health well-being.

**Applications and Model information:** Each article was reviewed to extract relevant information about the applications and models used. This included identifying the input modality, output modality, model, embodiment, open source status, and language for each application. We then categorized the applications and identified their target user groups.

**Validation methods and scales**: Each article was reviewed to extract relevant data and accurately document the key scales and metrics used in the studies. This involved identifying what each scale was measuring, such as usability, empathy, or coherence, and noting the specific items or questions that made up each scale. We distinguished between standardized scales, which are widely accepted and validated, and curated scales, which were adapted or created specifically for the studies. The score ranges and sample sizes for each scale were also recorded to help assess how reliable and applicable the findings are. References to the original studies were carefully documented to ensure the review is well-supported by existing research. Each section was also reviewed by two authors to maintain accuracy and consistency.

## Appendix C. Summary of LLM Platforms

| Article Title | Platforms |
|---|---|
| Appraising the performance of ChatGPT in psychiatry using 100 clinical case vignettes | OpenAI |
| The impact of prompt engineering in large language model performance: a psychiatric example | OpenAI |
| Can a Chatbot be Useful in Childhood Cancer Survivorship? Development of a Chatbot for Survivors of Childhood Cancer | - |
| Harnessing large language models' empathetic response generation capabilities for online mental health counselling support | - |
| Future of ADHD Care: Evaluating the Efficacy of ChatGPT in Therapy Enhancement | Raspberry PI |
| Investigating the Key Success Factors of Chatbot-Based Positive Psychology Intervention with Retrieval- and Generative Pre-Trained Transformer (GPT)-Based Chatbots | Baidu UNIT Dialogue Platform/OpenAI |
| An Empathic GPT-Based Chatbot to Talk About Mental Disorders With Spanish Teenagers | Telegram |
| Developing conversational Virtual Humans for social emotion elicitation based on large language models | Lab with virtual reality facilities |
| Supporting the Demand on Mental Health Services with AI-Based Conversational Large Language Models (LLMs) | - |
| MindfulDiary: Harnessing Large Language Model to Support Psychiatric Patients' Journaling | APP, no mention of iOS or Android |
| Facilitating Self-Guided Mental Health Interventions Through Human-Language Model Interaction: A Case Study of Cognitive Restructuring | Mental Health America |
| AI in relationship counselling: Evaluating ChatGPT's therapeutic capabilities in providing relationship advice | OpenAI |
| Cognitive Reframing of Negative Thoughts through Human-Language Model Interaction | Mental Health America |
| Loneliness and suicide mitigation for students using GPT3-enabled chatbots | Replika (iOS, Android, Oculus & Web) |
| Feasibility of combining spatial computing and AI for mental health support in anxiety and depression | - |
| Clinical decision support for bipolar depression using large language models | - |
| Exploring the Efficacy of Robotic Assistants with ChatGPT and Claude in Enhancing ADHD Therapy: Innovating Treatment Paradigms | Raspberry PI |

OpenAI uses a web platform, and by the time this manuscript is written, OpenAI models are also available on digital platforms including mobile phones and tablets; VR Lab deploys their system in a physical scenario within their lab; Telegram itself is used as a platform; Mental Health America uses a web platform; Replika is available on iOS, Android, Oculus (Meta's VR platform), and the web; Raspberry PI is a small single-board computer.

## Appendix D. Extracting and mapping evaluation constructs

We aligned the original constructs with those defined by Hua et al.'s[41] framework which scaffolds constructs into domains related to 1) safety, privacy, and fairness, 2) trustworthiness and usefulness, and 3) design and operations effectiveness. While Hua et al.'s framework is designed for general healthcare, and should be adapted for specific needs of mental health, it still offers a useful tool to understand the focus, goals, and overlap of constructs studied in the mental health space.

| Title | Evaluation Subject | Evaluation Method | Sample size | Scale Name | Original Construct | Mapped Second-Level Construct | Level |
|---|---|---|---|---|---|---|---|
| Appraising the performance of ChatGPT in psychiatry using 100 clinical case vignettes | LLM | Evaluated by generating responses to 100 clinical case vignettes in psychiatry, assessed by expert psychiatrists. | 100 vignette cases | - | Response Acceptability | Accessibility | 3 |
| Can a Chatbot be Useful in Childhood Cancer Survivorship? Development of a Chatbot for Survivors of Childhood Cancer | LLM | Evaluated by comparing the effectiveness of different models trained with and without domain-adaptive training, focusing on chatbot's accuracy and empathetic responses. | 46 samples | - | Empathy (in text-based mental health support) | Personalized engagement | 3 |
| Harnessing large language models' empathetic response generation capabilities for online mental health counselling suppor | LLM | Compared empathetic response generation capabilities using three empathy-related metrics against traditional empathetic dialogue systems and human responses. | 2545 conversations | - | Emotional Reactions: A helpseekers' attempt to address the emotional concerns of the person in distress | Personalized engagement | 3 |
| | | | | - | Interpretations: A help-seeker's attempt to restate the presenting problems of the | Personalized engagement | 3 |
| | | | | | person in distress | | |
| | | | | - | Exploration: The help-seeker's attempt to dive deeper into topics that the person in distress presents | Personalized engagement | 3 |
| Future of ADHD Care: Evaluating the Efficacy of ChatGPT in Therapy Enhancement | LLM | Evaluated by a panel of child ADHD therapy experts using the Delphi method, assessing effectiveness across therapeutic scenarios. | Not specified | - | Insight into Patient's Emotional State | Personalized engagement | 3 |
| | | | | - | Tailored and Personalized Responses | Personalized engagement | 3 |
| | | | | - | Overall Effectiveness as a Therapeutic Tool | Beneficence | 2 |
| | | | | - | Handling of Stress and Supporting Coping Mechanisms | Beneficence | 2 |
| | | | | - | Building Relationship and Trust | Accessibility | 3 |
| | | | | - | Sustaining Interest and Participation | Accessibility | 3 |
| | | | | - | Adaptation to Different Situations | Generalizability | 2 |
| | | | | - | Adaptability to Cultural and Sensory Differences | Generalizability | 2 |
| | | | | - | Handling of Sensitive Information | Privacy | 1 |
| Investigating the Key Success Factors of Chatbot-Based Positive | Users | Evaluated through three randomized controlled trials involving 326 participants, | 256 (Sub-study 1), 70 (Sub-study 2), 50 (Sub-study 3) participants | General Anxiety Disorder-7 (GAD-7) | Anxiety Severity | Beneficence | 2 |
| Psychology Intervention with Retrieval- and Generative Pre-Trained Transformer (GPT)- Based Chatbots | | focusing on effectiveness in mental health outcomes using Chatbot-Based Positive Psychology Interventions (Chat-PPIs). | 256 (Sub-study 1), 70 (Sub-study 2), 50 (Sub-study 3) participants | Positive and Negative Affect Schedule (PANAS) | Positive and Negative Affect | Beneficence | 2 |
| | | | 256 (Sub-study 1), 70 (Sub-study 2), 50 (Sub-study 3) participants | The Satisfaction With Life Scale (SWLS) | Life Satisfaction | Beneficence | 2 |
| | | | 256 (Sub-study 1), 70 (Sub-study 2) participants | The Subjective Vitality Scale (SVS) | Subjective Vitality | Beneficence | 2 |
| | | | 50 (Sub-study 3) participants | Psychological Wellbeing Scale (PWB) | Psychological Well-Being (6 dimensions) | Beneficence | 2 |
| An Empathic GPT-Based Chatbot to Talk About Mental Disorders With Spanish Teenagers | LLM | Evaluated by analyzing usage statistics, manual analysis of conversations, natural language processing techniques, and user feedback through an anonymous survey. | 44 participants | - | Chatbot Usability | Accessibility | 3 |
| | | | | - | User Engagement | Accessibility | 3 |
| | | | | - | Emotional Disclosure | Beneficence | 2 |
| | | | | - | User Satisfaction | Beneficence | 2 |
| Developing conversational Virtual Humans for social emotion elicitation based on large language models | LLM | Evaluated by measuring processing time, assessing human-computer interaction, and analyzing naturalness, realism, and emotional impact of | 64 participants | Patient Health Questionnaire (PHQ-9) | Depression | Beneficence | 2 |
| | | | | The State-Trait Anxiety Inventory (STAI) | Anxiety | Beneficence | 2 |
| | | | | - | Naturalness | Accessibility | 3 |
| | | | | - | Realism | Validity | 2 |

| | | | | virtual human interactions. | | The Self-Assessment Manikin (SAM) | Emotion (Valence and Arousal) | Accessibility | 3 |
|---|---|---|---|---|---|---|---|---|---|
| Supporting the Demand on Mental Health Services with AI-Based Conversational Large Language Models (LLMs) | LLM | Evaluated using intrinsic metrics like perplexity and extrinsic metrics including human assessments of response helpfulness, fluency, relevance, and logic. | 200 pairs | - | User Perceived Helpfulness | Accessibility | 3 |
| | | | | - | Response Fluency | Validity | 2 |
| | | | | - | Response Relevance | Cost-Effectiveness | 3 |
| | | | | - | Response Logic | Validity | 2 |
| MindfulDiary: Harnessing Large Language Model to Support Psychiatric Patients' Journaling | LLM | Evaluated through a four-week field study involving 28 patients with major depressive disorder, focusing on its effectiveness in facilitating journaling and enhancing clinical care. | 28 patients | Patient Health Questionnaire (PHQ-9) | Depression Severity | Beneficence | 2 |
| | | | | General Anxiety Disorder-7 (GAD-7) | Anxiety Severity | Beneficence | 2 |
| | | | | The Coping Scale | Coping Mechanisms | Beneficence | 2 |
| | | | | - | User Engagement | Accessibility | 3 |
| Facilitating Self-Guided Mental Health Interventions Through Human-Language Model Interaction: A Case Study of Cognitive Restructuring | Users | Evaluated through a large-scale, randomized study involving 15,531 participants, focusing on the impact on reducing emotional intensity and effectiveness of reframed thoughts. | 15531 participants | - | Reduction in Emotion Intensity | Beneficence | 2 |
| | | | | - | Reframe Relatability | Beneficence | 2 |
| | | | | - | Reframe Helpfulness | Beneficence | 2 |
| | | | | - | Reframe Memorability | Beneficence | 2 |
| | | | | - | Skill Learnability | Beneficence | 2 |
| AI in relationship counselling: Evaluating ChatGPT's therapeutic | LLM | Evaluated based on technical metrics like error rate, linguistic | 20 participants | - | Usability | Accessibility | 3 |
| | | | | - | Technical Issues | Reliability | 2 |
| | | | | - | Task Completion Rate | Validity | 2 |
| capabilities in providing relationship advice | | accuracy, and therapeutic quality indicators, analyzed through content and reflexive thematic analysis of participant interviews. | | - | Dialogue Efficiency | Cost-effectiveness | 3 |
| | | | | - | Dialogue Handling | Personalized Engagement | 3 |
| | | | | - | Context Awareness | Generalizability | 2 |
| | | | | - | Error Management | Reliability | 2 |
| | | | | - | Appropriateness of Response | Accessibility | 3 |
| | | | | - | Comprehensibility | Validity | 2 |
| | | | | - | Realism | Personalized Engagement | 3 |
| | | | | - | Empathy | Personalized Engagement | 3 |
| | | | | - | Repetitiveness of Response | Reliability | 2 |
| | | | | - | Linguistic Accuracy | Validity | 2 |
| | | | | - | Chatbot's Understanding of Response | Validity | 2 |
| | | | | - | Reflection | Personalized Engagement | 3 |
| | | | | - | Validation | Personalized Engagement | 3 |
| | | | | - | Therapeutic Questioning | Validity | 2 |
| | | | | - | Unrushed Approach | Personalized Engagement | 3 |
| | | | | - | Addressing Safety Concerns | Safety | 1 |
| | | | | - | Collaborative Solutions | Beneficence | 2 |
| | | | | - | Response Length | Cost-effectiveness | 3 |
| | | | | - | Overall Sense of Flow and Coherence | Personalized Engagement | 3 |
| Cognitive Reframing of Negative Thoughts through Human-Language Model Interaction | Users | Evaluated through a randomized field study on a large mental health platform with 2,067 participants, assessing the relatability, helpfulness, | 2067 participants | - | Relatability | Accessibility | 3 |
| | | | | - | Helpfulness | Beneficence | 2 |
| | | | | - | Memorability | Personalized engagement | 3 |
| | | | | | and memorability of LLM-generated reframed thoughts. | | |
| Loneliness and suicide mitigation for students using GPT3-enabled chatbots | Users | Evaluated through a survey of 1006 student users of Replika ISA, analyzing loneliness, perceived social support, use patterns, and beliefs, combined with qualitative coding and statistical analysis. | 1006 participants | De Jong Gierveld Loneliness Scale | Loneliness (overall, emotional, social) | Beneficence | 2 |
| | | | | Interpersonal Support Evaluation List (ISEL) | Perceived social support | Beneficence | 2 |
| Feasibility of combining spatial computing and AI for mental health support in anxiety and depression | Users | Evaluated through qualitative analysis of therapy transcripts, focusing on constructs like therapeutic alliance, empathy, emotional engagement, and effectiveness of VR environment. | 14 participants | - | Therapeutic Alliance (Perceived connection and trust between participants and the system) | Accessibility | 3 |
| | | | | - | Empathy | Personalized engagement | 3 |
| | | | | - | Emotional Engagement | Personalized engagement | 3 |
| | | | | - | Effectiveness of VR Environment (Impact of the VR environment on participants' relaxation and emotional comfort) | Accessibility | 3 |
| | | | | - | CBT Compliance | Beneficence | 2 |
| | | | | - | Usability and User Experience | Accessibility | 3 |

| Clinical decision support for bipolar depression using large language models | LLM | Evaluated by comparing the model's ability to select optimal next-step pharmacotherapy for bipolar depression against expert consensus. | 50 vignettes | - | Treatment Appropriateness | Beneficence | 2 |
|---|---|---|---|---|---|---|---|
| Exploring the Efficacy of Robotic Assistants with ChatGPT and Claude in Enhancing ADHD Therapy | LLM | Evaluated in robotic-assisted ADHD therapy sessions using both technical and clinical evaluations, including therapist feedback using the Delphi method. | Not specified | - | Facilitation of Safe Emotional Expression | Safety | 1 |
| | | | | - | Validation of Patient's Experiences and Emotions | Personalized engagement | 3 |
| | | | | - | Consistency and Appropriateness of Empathy | Personalized engagement | 3 |
| | | | | - | Empathetic Response to Emotional Indicators | Personalized engagement | 3 |
| | | | | - | Insight into Patient's Emotional State | Validity | 2 |
| | | | | - | Clarity and Comprehensibility of Communication | Cost-effectiveness | 3 |
| | | | | - | Coherence and Relevance in Conversation | Reliability | 2 |
| | | | | - | Clarity and Conciseness of Information Provided | Cost-effectiveness | 3 |
| | | | | - | Handling Misunderstandings | Reliability | 2 |
| | | | | - | Multilingual Interaction Handling | Accessibility | 3 |
| | | | | - | Positive Session Atmosphere | Beneficence | 2 |
| | | | | - | Encouragement of Autonomy and Self-expression | Beneficence | 2 |
| | | | | - | Sustaining Patient Interest | Personalized engagement | 3 |

| | | | | | - | Promotion of Active Participation | Personalized engagement | 3 |
|---|---|---|---|---|---|---|---|---|
| | | | | | - | Engagement Level in Therapy Sessions | Personalized engagement | 3 |
| | | | | | - | Engaging and Motivational Language Usage | Personalized engagement | 3 |
| | | | | | - | Adjustment Based on Feedback | Personalized engagement | 3 |
| | | | | | - | Flexibility in Conversational Style | Accessibility | 3 |
| | | | | | - | Ability to Redirect Conversation | Personalized engagement | 3 |
| | | | | | - | Response to Novel or Unexpected Inputs | Generalizability | 2 |
| | | | | | - | Adaptability to Changing Conversation Dynamics | Personalized engagement | 3 |
| | | | | | - | Respect for Patient's Boundaries | Beneficience | 2 |
| | | | | | - | Creation of a Safe Environment | Beneficience | 2 |
| | | | | | - | Building Trust with Patient | Personalized engagement | 3 |
| | | | | | - | Compatibility with Various Therapeutic Modalities | Reliability | 2 |
| | | | | | - | Potential for Future Applications | - | |
| | | | | | - | Recommendation for Clinical Use | Cost-effectiveness | 3 |
| | | | | | - | Suitability for Diverse Patient Groups | Generalizability | 2 |
| | | | | | - | Meaningful Contributions to Therapy | Beneficience | 2 |
| | | | | | - | Overall Effectiveness as a Therapeutic Tool | Beneficience | 2 |

| | | | | - | Performance and Accuracy | Validity | 2 |
|---|---|---|---|---|---|---|---|
| | | | | - | Response Time | Cost-effectiveness | 3 |
| | | | | - | Understanding and Coherence | Validity | 2 |
| | | | | - | Safety and Bias | Safety; Fairness and bias management | 1 |
| | | | | - | Customization and Flexibility | Accessibility | 3 |
| | | | | - | Integration Ease | Cost-effectiveness | 3 |
| | | | | - | Innovation | - | |
| | | | | - | Multilingual Support | Accessibility | 3 |